\documentclass[11pt]{article}
\usepackage[]{acl}

\usepackage{times,latexsym}
\usepackage{url}
\usepackage[T1]{fontenc}
\usepackage{microtype}
\usepackage{multirow}
\usepackage{comment}
\usepackage{multicol}
\usepackage{lscape}
\usepackage{graphicx}
\usepackage{enumitem}
\usepackage{xspace,mfirstuc,tabulary}
\setitemize{noitemsep,topsep=0pt,parsep=0pt,partopsep=0pt}
\usepackage{titlesec}
\usepackage{booktabs}

\usepackage[T1]{fontenc}
\usepackage[utf8]{inputenc}

\usepackage{microtype}

\title{Re-contextualizing Fairness in NLP: The Case of India}

\author{
Shaily Bhatt \\ Google Research \\ shailybhatt@google.com \And
Sunipa Dev \\ Google Research \\ sunipadev@google.com \And
Partha Talukdar \\ Google Research \\ partha@google.com \AND
Shachi Dave* \\ Google Research \\ shachi@google.com \And
Vinodkumar Prabhakaran* \\ Google Research \\ vinodkpg@google.com}

\begin{document}
\maketitle
\begin{abstract}

Recent research has revealed undesirable biases in NLP data and models. However, these efforts focus on social disparities in West, and are not directly portable to other geo-cultural contexts. In this paper, we focus on NLP fairness in the context of India.
% , a country of 1.4 billion people with a fast-growing NLP ecosystem. 
We start with a brief account of the prominent axes of social disparities in India. 
% : \textit{Region}, \textit{Caste}, \textit{Gender}, \textit{Religion}, \textit{Ability}, and \textit{Gender identity \& sexual orientation}. 
We build 
resources
% \href{https://github.com/google-research-datasets/nlp-fairness-for-india}{resources} 
for fairness evaluation in the Indian context 
% in NLP data and models along some of these axes.
and use them to demonstrate prediction biases along some of the axes. We then delve deeper into social stereotypes for Region and Religion, demonstrating its prevalence in corpora and models.
Finally, we outline a holistic research agenda to re-contextualize NLP fairness research for the Indian context, accounting for Indian \textit{societal context}, bridging \textit{technological} gaps in NLP capabilities and resources, and adapting to Indian cultural \textit{values}. While we focus on India, this framework can be generalized to other geo-cultural contexts.

\end{abstract}

\section{Introduction}
\label{sec1_intro}
While Natural Language Processing (NLP) has seen impressive advancements recently \cite{devlin2018bert, raffel2019exploring, brown2020language, chowdhery2022palm},
% , and wide-spread adoption in applications like content moderation \cite{akiwowo2020proceedings,oshikawa2020survey}, information retrieval \cite{tiwary2008natural}, digital assistants \cite{maedche2019ai}, etc. 
% \& domains like employment \cite{adamczyk2021automation}, healthcare \cite{gu2021domain}, finance \cite{mishev2020evaluation}, \& education \cite{smith2020computer}.
it has also been demonstrated that language technologies may capture, propagate, and amplify societal biases
% \footnote{We use \textit{stereotype} to refer to perceptions (often untrue) associated with social or personal identities. We intentionally do not try to define ``harm'' here as we believe that what stereotypes are ``harmful”, to whom, and how are questions that require deeper engagements with communities (See \S\ref{sec6_discussion}), which is out of the scope of this work in this work} 
\cite{blodgett2020language}. 
Although NLP is adopted globally, most studies on assessing and mitigating biases are in the Western context,\footnote{We use \textit{Western} or \textit{the West} to refer to the regions, nations \& states consisting of Europe, the U.S., Canada, and Australasia, and their shared norms, values, customs, religious beliefs, \& political systems \cite{kurth2003western}.} focusing on axes of disparities in the West, relying on Western data and justice norms, and are not directly portable to non-Western contexts \cite{sambasivan2021re}. 

This is especially troubling for India,
% India is 
a pluralistic nation of 1.4 billion people, 
% having multiple religions, ethnicities, and cultural histories, 
% India is also seeing 
with 
% an increased interest and 
fast-growing investments in NLP from the government and the private sector.\footnote{In government (\url{https://bhashini.gov.in}) and private sector (\url{https://tinyurl.com/indiaai-top-nlp-startups}, \url{https://tinyurl.com/google-idf-language}).}
% \footnote{\href{https://bhashini.gov.in}{Bhashini}, \href{https://bhashini.gov.in}{NLP Startups in India}, \href{https://bhashini.gov.in}{Google IDF}}
% \url{tinyurl.com/toi-nlp},
% % \url{tinyurl.com/progress-in-nlp-businesstoday},
% \url{bhashini.gov.in/en/},
% % \url{tinyurl.com/google-idf-language},
% \url{tinyurl.com/indiaai-top-nlp-startups}
There is commendable recent work on fairness in NLP models for Indian languages such as Hindi, Bengali, and Telugu \cite{pujari2019debiasing, malik2021socially, gupta2021evaluating}.
% fairness research for the Indian context. This requires accounting for the various relevant axes of social disparities in Indian society, their proxies in language data, and the (lack of) availability of resources that enable fairness evaluations and mitigation.
But, for a nation with many religions, ethnicities, and cultures, re-contextualizing NLP fairness needs to account for the various axes of social disparities in the Indian society, their proxies in language data, the disparate NLP capabilities in Indian languages, and the (lack of) resources for bias evaluation.

\citet{sambasivan2021re} proposed a research agenda for AI fairness for India based on interviews of 36 experts on Indian society and technology.
% they do not provide any empirical analysis of biases reflected in NLP data or models. 
In this paper, we build on their work with a focus on NLP. 
% develop an NLP specific research agenda for fairness in the Indian context. 
We start with a brief discussion on the major axes of social disparities in India 
% \& their manifestation in language data 
(\S\ref{sec3_axes}). 
% We distinguish axes that are fairly unique to India, such as \textit{Caste} and \textit{Region}, from the globally salient axes such as \textit{Gender}, \textit{Religion}, \textit{Ability}, and \textit{Gender identity and sexual orientation} that may have unique manifestations in the Indian context. 
We then discuss the proxies of some of these axes in language and empirically demonstrate prediction biases around these proxies in NLP models (\S\ref{sec4_proxies}). We then delve deeper into stereotypes along the axes of \textit{Region} and \textit{Religion}, demonstrating their prevalence in data and models (\S\ref{sec5_india_stereotypes}). 
Finally, we build on these empirical demonstrations to propose an overarching research agenda along the \textit{societal}, \textit{technological}, and \textit{value alignment} aspects important to formulating fairness research for the Indian context (\S\ref{sec6_discussion}). While we focus on India in this paper, our framework can be adapted to re-contextualize fairness research for other geo-cultural contexts. 
% We focus on the \textit{societal context} for empirical demonstrations and then provide an overarching research agenda additionally covering \textit{technological} and \textit{value alignment} aspects important to formulating fairness research for the Indian context (\S\ref{sec6_discussion}). While we focus on `India', our framework can be adapted to re-contextualize fairness research for other geo-cultural contexts.
% Finally, we identify a set of major challenges in NLP Fairness in the Indian context, and put forth a research agenda to address these.

To summarize, our main contributions are: 
(1) an overarching research agenda for NLP fairness in the Indian context accounting for societal, technological, and value aspects;
(2) resources (curated and created) for enabling fairness evaluations in the Indian context available;\footnote{\url{https://www.github.com/google-research-datasets/nlp-fairness-for-india}}
and (3) empirical demonstrations of prediction biases and over-prevalence of social stereotypes in data and models.

\section{Related Work}
\label{sec2_background}
% \begin{figure*}[t]
% \centering
% \includegraphics[width=\linewidth]{Re-contextualizing_Fairness_in_NLP_The_Case_of_India/images/bindi_agenda.png}
% % \caption{States and union territories of India. (Source: \url{https://indiamaps.gov.in/soiapp/}); Axes of Social Disparities and \% minority group populations; Proxies in language that \textit{may} reveal sub-group identity.
% \caption{A holistic research agenda for NLP Fairness in the Indian context: accounting for societal disparities in India (Section~\ref{sec3_axes}-\ref{sec5_india_stereotypes}), bridging technological gaps in NLP capabilities/resources, and adapting fairness interventions to align with local values and norms (Section~\ref{sec6_discussion}). (Map source: \url{https://indiamaps.gov.in/soiapp/})
% % \caption{Social disparities and proxies in language relevant for NLP Fairness in India. (Map source: \url{https://indiamaps.gov.in/soiapp/})
% % NLP Fairness in the Indian Context. \vp{Only a placeholder figure; Feel free to design your own at \url{go/bindi-paper-figures-deck}; Objective: a pictorial representation of the agenda} \vp{Map source: \url{https://worldmapwithcountries.net/2020/03/12/india-map-with-states} to confirm copyright}
% }
% \label{fig:Indiaframework}
% \end{figure*}

% \subsection{Bias in NLP}
% \paragraph{Societal Biases in NLP: }
Research on undesirable biases has been a growing priority in NLP  \cite{caliskan2017semantics,blodgett2020language,sheng2021societal,ghosh2021detecting}. Social biases are shown to be baked into pretrained language models \cite{bender2021dangers} and models for downstream tasks such as sentiment analysis \cite{kiritchenko2018examining} and toxicity detection \cite{sap2019risk}. While the majority of NLP fairness research focuses on gender \cite{bolukbasi2016man,sun2019mitigating,zhao2017men} and racial biases \cite{sap2019risk,davidson2019racial,manzini2019black}, other axes of disparities such as ability \cite{hutchinson2020social}, age \cite{diaz2018addressing}, and sexual orientation \cite{garg2019counterfactual} have also gotten some attention. However, the majority of this research is framed in and for the Western context, relying on data and values reflecting the West \cite{sambasivan2021re}.

Recently, fairness research in NLP has also been expanded to non-English
% There is recent expanding fairness research to 
languages such as Arabic \cite{lauscher-etal-2020-araweat}, Japanese \cite{Takeshita2020CanEM}, Hindi, Bengali, and Telugu \cite{pujari2019debiasing,malik2021socially}.
% , Dutch \citet{chavez-mulsa-spanakis-2020-evaluating}, German and French \citet{kurpicz2020cultural}. 
Evidence of 
% Researchers have also demonstrated 
cultural biases for different countries have also been recorded \cite{ghosh2021detecting} in LMs. Our work adds to this line of research. Building on \citet{sambasivan2021re}, we take a more holistic approach towards NLP fairness in the specific geo-cultural context of India.
More specifically, we re-frame the agenda they proposed along ``re-contextualising data and model fairness; empowering communities by participatory action; and enabling an ecosystem for meaningful fairness'' with an NLP-centric lens.

\section{Axes of Disparities}
\label{sec3_axes}

Identifying prominent axes of disparities is the first step in laying out a holistic NLP fairness research agenda for the Indian context. We follow \citet{sambasivan2021re} who identify the major axes of potential ML (un)fairness (Table 1 of \citet{sambasivan2021re}), and include \textit{Region}, \textit{Caste}, \textit{Gender}, \textit{Religion}, \textit{Ability}, and \textit{Gender Identity and Sexual Orientation}.\footnote{\citet{sambasivan2021re} include \textit{Class} as an axis, however we see class as an attribute that cuts across multiple axes, rather than as an immutable characteristic.}
We further group them into globally salient axes (such as \textit{Gender} and \textit{Religion}) with local manifestations (such as different religions - for example, \textit{Jainism}) and axes that are unique and/or specific to India (such as \textit{Region} and \textit{Caste}).

Further, amplified social biases may be faced by those with overlapping categories of marginalized groups. We do not focus on this \textit{Intersectionality} here and leave discussion about it to Section \ref{sec6_discussion}.

\subsection{India-specific axes}

\paragraph{Region:} Region as an axis can manifest globally (for example as nationality), but here we predominantly focus on the ethnicity associated with geographic regions of India
and hence categorize it as India-specific. 
While the census does not recognise racial or ethnic groups,\footnote{\url{https://www.censusindia.gov.in/}} India is home to many ethno-lingusitic groups with diverse cultures and traditions.\footnote{\url{https://tinyurl.com/SA-ethnic-groups}} 
% While many of these ethnolinguistic groups do cross state boundaries, %, since state borders were originally drawn based on linguistic divisions, 
Most states in India comprise a dominant ethno-lingusitic group (such as \textit{Haryanvis} in \textit{Haryana}, \textit{Goans} in \textit{Goa}). 
Early research has documented various stereotypes for regional subgroups \cite{borude1966linguistic,de1977regional}. \newcite{de1977regional} reported that students from a college in Mumbai ascribed traits such as crooked to Andhraites, cunning to Kannadigas, and brave to Punjabis, observing that South Indians were ascribed ``unfavorable'' traits more frequently. Disparities and stereotypes also exist in India at broader regional levels (for example, negative stereotypes and rampant discrimination has been documented against North-East Indians \cite{mcduie2012northeast,haokip2021chinky}), and groups belonging to smaller regions within or across states (like Konkani in parts of Goa, Maharashtra, and Karnataka). 

\paragraph{Caste:} Caste is an inherited hierachical social identity, that has been basis of historical marginalization.
% ``that can determine all aspects of one's life opportunities, including personal rights, choices, freedom, dignity, access to capital and effective political participation in caste-affected societies'' \cite{shanmugavelan2018everyday}. 
Despite the intended eradication of 
% varied alterations over the centuries \cite{deshpande2010history}, and 
% eradication of 
caste-based discrimination envisioned decades ago \cite{ambedkar2014annihilation}, lower rungs of the caste hierarchy continue to have low literacy rates, misrepresentation, poverty, low technology access, and exclusion in language data \cite{deshpande2011grammar,kamath2018untouchable,krishna2019does}.\footnote{\url{https://tinyurl.com/oxfamindia-caste}} Caste-based prejudices have been documented in matrimonial ads \cite{rajadesingan2019smart} and social media \cite{vaghela2021birds}. \newcite{fonseca2019caste} found that news coverage of ``lower caste'' groups were focus excessively on prejudice, violence, and conflict, and ignore other aspects of their life and identity.  
% Not much work has studied how these stereotypes may be captured and propagated through models trained on internet.

\subsection{Global axes in the Indian context}

\paragraph{Gender:} Although gender is a prominent axis of disparity across the globe, the specifics of how gender manifests in society (and hence, in data) varies greatly across geo-cultural contexts \cite{kurian2020sex}.
% \footnote{\url{https://tinyurl.com/gender-geocultural-diff}} 
% Hence, studies on gender bias in NLP performed using data from the West may need to be recontextualized to the Indian context in order for them to be effective. This
Re-contextualization of the gender axis needs to account for India-specific gender stereotypes and the structural 
% factors that shape the 
disparities in engagement of women in society. 
For example women in India are 58\% less likely to connect to mobile Internet then men \cite{sambasivan2019toward}, have literacy rate of 65\% compared to 85\% for men, and 21\% labor force participation compared to 76\% for men.\footnote{\url{https://tiny.cc/labor-gender-in}} 
% Studies have also documented how 
Gender roles and stereotypes in India vary from the West \cite{sethi1984sex,leingpibul2006cross} and so do their potrayal in media \cite{griffin1994gender,khairullah2009cross,das2011gender}. 

\begin{comment}
% \vp{Discuss how gender manifestations may be similar and different between India and the West. Cite social science studies that has documented this, and/or statistics from various governmental sources.}
% For instance, in case of gender, women in India have much lower literacy rates, work-force representation, mobility, and socio-political power in the patriarchy as compared to women in the West. %where the literacy rates and workforce representations of men and women are roughly equal. 
\end{comment}

\paragraph{Religion:} Religious biases have been studied in NLP \cite{dev2020measuring, stereoset-paper, anti-muslim-bias-gpt3}, however the social disparities and stereotypes about various religious groups differ significantly in India from the West,~\cite{malik2021socially}. 
For example, Christianity (typically a majority religion in the West) is a minority religion (2.3\% of the population) in India, along with Sikhism (1.9\%), Buddhism (0.8\%), and Jainism (0.4\%). 
% While religious minorities often occupy marginalized positions in Indian society along social, economic and educational aspects \cite{basant2007social}, not many studies have looked into how religious stereotypes in India are reflected in NLP. 

\begin{comment}
% \vp{We need to do more lit survey here; While I see some work, I don't see a lot. But saying there is no work might be an over-statement}

% .\footnote{\url{https://tinyurl.com/pew-indian-christianity}} 
% \vp{Discuss how religion biases may be similar and different between India and the West. Discuss how Christians are a minority in India.}
% Uniqueness in subgroups is seen in religion where Western fairness formulations of religion typically focus on Christianity, Islam, and Judaism, the Indian population consists of a majority of Hindus. Christianity and Islam are minority religions, along with other (prominent but minority) religions such as Jainism, Zoroastrianism, Buddhism, and Sikhism.
\end{comment}

\paragraph{Ability:} Awareness about (dis)ability is relatively recent in India \cite{ghosh2016interrogating,ghai2019rethinking}. Representation of disability in social discourse and the barriers it poses are significantly different for India than the West \cite{chaudhry2005rethinking,johnstone2017disability}. For example people with disabilities are often abandoned at birth or socially segregated \cite{kumar2012disability} due to being seen as deceitful, unable to progress to adulthood, and dependent on charity and pity \cite{ghai2002disabled}. Disability is often mocked, portrayed as a punishment, and heteronormative narratives of `fixing' disability are prevalent in Indian cinema \cite{sawhnettracing}. 
% \vp{Discuss how disability biases may be similar and different between India and the West.} 

\paragraph{Gender Identity and Sexual Orientation:} Discourse around gender identity and sexual orientation has historically been largely absent from the Indian public discourse \cite{abraham1998homosexuality}. While India reflects the growing positive attitude towards LGBTQ+ issues \cite{anand2016attitude} along with the recent decriminalisation of homosexuality \cite{tamang2020section}, there still exist challenges to acceptance and visibility. Furthermore, understanding LGBTQ+ related biases in the Indian context needs engagement with the social situatedness of groups like the \textit{hijra} community, a socially outcast intersex and transgender community.

\section{Proxies of Axes and Predictive Disparities}
\label{sec4_proxies}
Bias evaluation in NLP relies on proxies of subgroups in language, such as identity terms and personal names,
% to conduct evaluation, 
% as they may reveal 
to reveal the undesirable associations present in models and data \cite{caliskan2017semantics,maudslay2019s}. 
% These proxies include identity terms (i.e, terms used to identify various subgroups) \cite{garg2019counterfactual,hutchinson2020social}, and personal names with identity associations \cite{caliskan2017semantics,maudslay2019s}. 
In the Indian context, we identify three major kinds of proxies: \textit{identity terms}, \textit{personal names}, and \textit{dialectal features}.

Using such proxies however poses unique challenges in the Indian context. For example, there are thousands of caste identities and hundreds of ethno-linguistic regional identities that are not codified in any authoritative sources. Similarly, there do not exist any large resources that provide subgroup associations for personal names, such as the US Census data (for race) or SSA data (for gender) in the West. Building exhaustive resources to capture such fine-grained social groups is outside the scope of this paper. However, in this section we curate identity terms and personal names with prototypical identity associations. We adopt a black-box evaluation strategy to demonstrate predictive biases in standard NLP pipelines/models and also demonstrate the utility of India-specific resources. Finally we note that these resources and studies are meant to be demonstrative, not exhaustive. 
% will need to be expanded in future work for comprehensiveness.
% 
% In the rest of this section, we present these resources and use them to inspect NLP model biases along various axes outlined in Section~\ref{sec3_axes}. These studies are not meant to be exhaustive, but to demonstrate the utility of India-specific resources.

\subsection{Identity Terms}
\label{identity_terms}

We curated lists of India-specific identity terms along three different axes:

\begin{itemize}
    \item \textit{Region}:
    % a list of identity terms based on 
    demonyms for states \& union territories like \textit{Kashmiri}, \textit{Andamanese}.\footnote{\url{https://tinyurl.com/wiki-in-regions}}
    \item \textit{Caste}: frequently used terms-\footnote{Broad (and overlapping) categories, not caste names.} \textit{Brahmin}, \textit{Kshatriya}, \textit{Vaishya}, \textit{Shudra}, \textit{Dalit}, \textit{SC/ST} (Scheduled Castes/Scheduled Tribes), \textit{OBC} (Other Backward Classes).
    \item \textit{Religion}: terms for populous religions- \textit{Hindu}, \textit{Muslim}, \textit{Christian}, \textit{Sikh}, \textit{Buddhist}, \textit{Jain}.
\end{itemize}

\noindent We now demonstrate biases in the default HuggingFace sentiment pipeline which is DistilBERT-base-uncased \cite{sanh2019distilbert} fine-tuned on the SST-2  \cite{socher2013recursive}.\footnote{\url{https://tinyurl.com/hf-sentiment}.
% We use the default model from HuggingFace for ease of replicability.
} 
We perform perturbation sensitivity analysis \cite{prabhakaran2019perturbation} that reveals biases by counterfactual replacement of terms of same semantic category in natural sentences. For example, the sentence ``Gujarati people love food.'' is perturbed with regional identity terms leading to sentences like ``Kashmiri people love food'', ``Andamanese people love food'' etc. We report the normalized shift in sentiment scores for these perturbed sentences, essentially demonstrating the degree to which the scores are affected by the identity term present in the sentence.

For this analysis, we extract sentences in which an identity term occurs from IndicCorp-en \cite{kunchukuttan2020ai4bharat}, and randomly select equal number of sentences for every identity term 
% balance the sample size across identity terms 
to prevent the topical content from being biased towards any subgroup. We extract 10, 150, \& 200 sentences, totalling in 357, 1050, and 1200 sentences along region (some region terms had less than 10 sentences), caste, and religion respectively. 
% We find sentiment scores to shift on perturbation with different identity terms (Figure~\ref{fig:caste_religion_psa} shows the sentiment score shifts for different regional identities and Figure~\ref{fig:caste_religion_psa} shows it for caste and religious identities). In particular: for region 

\begin{figure}[]
\centering
\includegraphics[width=.6\linewidth]{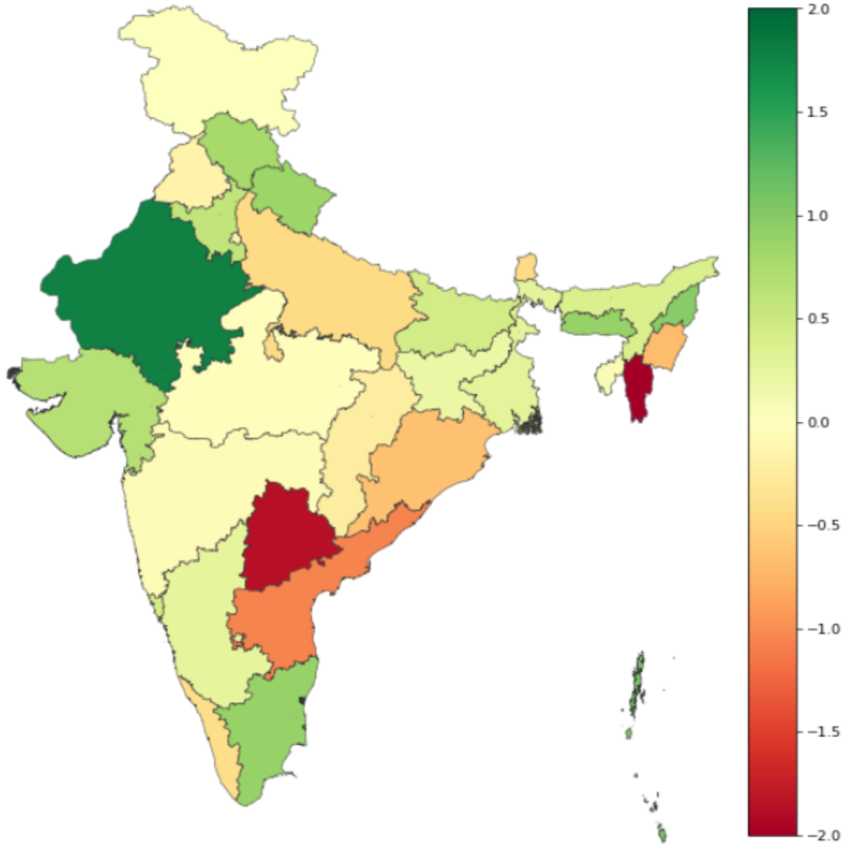}
\caption{Relative sentiment score shift when regional identity terms are perturbed showing negative (e.g., \textit{Mizoram}) and positive (e.g., \textit{Rajasthan}) associations. 
% \vp{Shaily, there were many missing states in the file you sent; could you please double check and send a fuller dataset? No need to sample}
}
\label{fig:region_sentiment_hf}
\end{figure}

\begin{figure}[]
\centering
\includegraphics[width=.9\linewidth]{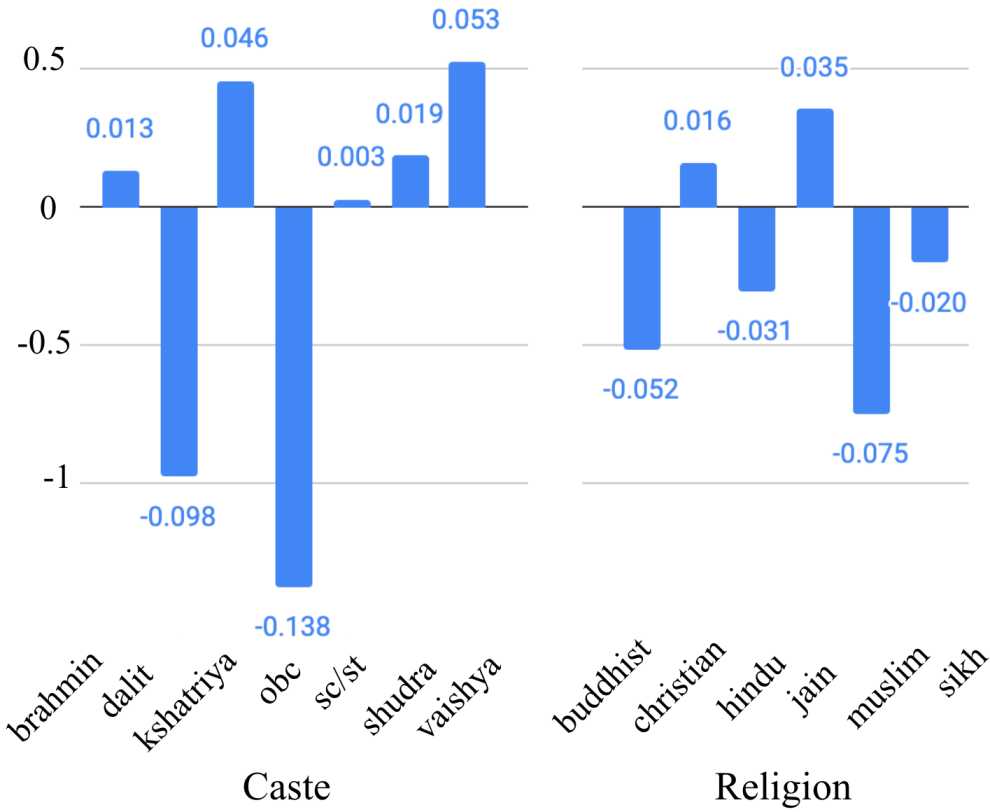}
\caption{Relative sentiment score shift when caste and religious identities are perturbed showing negative associations with marginalized groups (e.g. \textit{obc}, \textit{muslim}).}
\label{fig:caste_religion_psa}
\end{figure}

Figure~\ref{fig:region_sentiment_hf} shows the shift in scores for regional identities. We find \textit{Mizoram} and \textit{Telangana} have among the most negative score shifts, while \textit{Rajasthan} and \textit{Gujarat} had among the most positive association. Figure~\ref{fig:caste_religion_psa} shows the relative shift for caste and religion. For caste, the model had significant negative association towards the terms \textit{obc} and \textit{dalit}, both of which represent historically marginalized groups; and for religion, we find negative association towards the terms \textit{muslim} and \textit{hindu}, while \textit{jain} and \textit{christian} have positive associations.

\subsection{Personal Names}
\label{personal_names}

% As \citet{sambasivan2021re} noted,  
Personal names \textit{can be} strong proxies for various socio-demographic identity groups in India, including gender, religion, caste, and regional ethnolinguistic identities \cite{sambasivan2021re}.
% \paragraph{Curation of Personal Names}
We curate a list of Indian first names with prototypical binary gender association
% , and last names that have prototypical association with ethno-linguistic regional groups
. We build this list 
% (with the prototypical gender association) 
by querying the MediaWiki API using a seed list of Wikipedia category pages listing Indian names.\footnote{\url{https://tinyurl.com/wiki-indian-names}}  

\begin{comment}
% We only use personal names associated with binary gender cisgender categories because there are no such prototypical name associations with other genders. The number of personal names curated for each sub-group identity term is indicated in Appendix~\ref{app-a:idterms} and we will release the full list with the paper.

% \paragraph{DisCo Analysis for Gendered Correlations:}
% \label{disco-analyis}
\end{comment}

We now perform analysis of gendered correlation in pretrained models using the DisCo metric \cite{filbert-paper} which measures if the predictions of a language model have disproportionate association to a particular gender. Following \citet{filbert-paper}, we perform slot filling using a set of templates and names, and record the number of candidate words generated by the language model having statistically significant association with a gender, averaged over the number of templates. A higher value for DisCo metric means more associations. 
% In particular, we use  
% to evaluate gendered correlations. 
% The metric measures if the predictions of a language model have disproportionate association to a particular gender.
% It 
We analyze two language models: MuRIL \cite{khanuja2021muril} and multilingual BERT (mBERT) \cite{devlin2018bert}. MuRIL uses the same architecture as mBERT, but is trained on more data derived from the Indian context, and significantly outperforms mBERT on multiple benchmark tasks for Indian languages, including ~20\% improvement in NER. 

We calculate DisCo metric in two ways: (1) using a list of 300 American male and female names (such as, \textit{Mary}, \textit{John}) and (2) using 300 Indian male and female names (such as, \textit{Rahul}, \textit{Pooja}). 

\begin{figure}[]
\centering
\includegraphics[width=.75\linewidth]{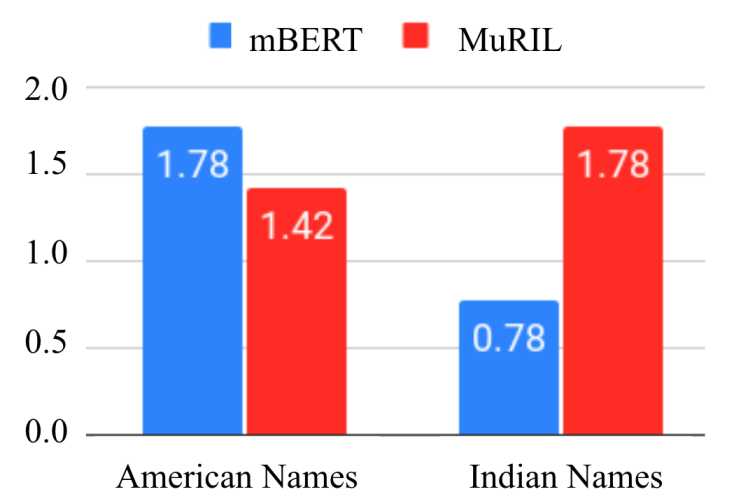}
\caption{DiSCO metric (higher value means more gendered correlations) mBERT and MuRIL 
% calculated with American (left) and Indian (right) names.}
% showing that MuRIL encodes more bias around Indian names compared to American Names. 
}
\label{fig:bindi_filbert}
\end{figure}

Results in figure~\ref{fig:bindi_filbert} leads to 2 observations. First, in line \citet{filbert-paper}, gender bias is encoded for personal names in the Indian context. Second, India-specific resources are critical to bias evaluation. This is because, using American names, it appears that MuRIL has a lesser amount of bias than mBERT. However, using Indian names reveals that while MuRIL learned to detect names better (i.e., improved NER performance), it also learned more stereotypical associations around those names.

\begin{comment}
% As shown in Figure~\ref{fig:bindi_filbert}, if we calculate the DisCo metric using a set of 300 American names with strong gender association, it appears that MuRIL has a lesser amount of gender bias than mBERT. 
% However, replacing the set of names with 300 Indian names with gender association, reveals that while MuRIL learned to detect more Indian names mentioned in text (i.e., improved NER performance), it also learned more and stronger stereotypical associations around those names. 

% This observation has two fold implications. First, in line with findings of \cite{filbert-paper}, bias is also encoded along personal names even in the Indian context (i.e gendered correlations encoded for Indian male and female names). And secondly, it demonstrates that India-specific resources are critical to bias evaluation as western resources (like western name lists) may not give accurate estimates of bias encoded in Indic models.

% \subsubsection{Name Perturbation Analysis}
% \sjb{TODO(Vinod/Sunipa/Shaily)}
% \vp{Present analysis using perturbation on personal names with gender associations. TODO. Shaily, could you please share the names list with me?}
\end{comment}

\subsection{Dialectal Features}
\label{dialect}

% Another major proxy for various subgroups is the 
Presence of dialectal features
% in language use 
% is a noth. Dialects 
is often associated with demographic subgroups (like socio-economic class \cite{bernstein1960language,kroch1978toward}), and hence can act as a proxy for many axes.
Dialects are not monolithic; distinctions are often captured by the presence, absence, and frequency
of many features (such as, \textit{article omission}) \cite{demszky2021learning}.
% Recognizing dialect features in text at scale is an open research problem. 
For this study, we use the minimal pairs dataset built by \cite{demszky2021learning} with 266 sentences annotated with presence of 22 morpho-syntactic dialectal features prevalent in Indian English. For each sentence with a dialect feature, the dataset also contains an equivalent sentence without the feature; effectively functioning as a counterfactual dataset for dialect features. We run this dataset through the sentiment model described earlier, and assess its sensitivity to the presence of dialect features. 
% This is similar to the analysis performed by \cite{checklist_paper} demonstrating robustness failures in NLP models.

% we used the minimal pairs dataset curated by  \cite{demszky2021learning}. This dataset contains 266 pairs of sentences spanning 22 dialectal features of annotated with dialectal features prevalent in Indian English, released by \cite{demszky2021learning}. The dataset contains 266 sentences annotated with presence of 22 different dialectal features. We run these sentences through the sentiment model and report whether there is a shift in the prediction scores in response to the presence of dialect features. This is similar to the analysis performed by \cite{checklist_paper} demonstrating robustness failures in NLP models.

\begin{figure}[]
\centering
\includegraphics[width=.9\linewidth]{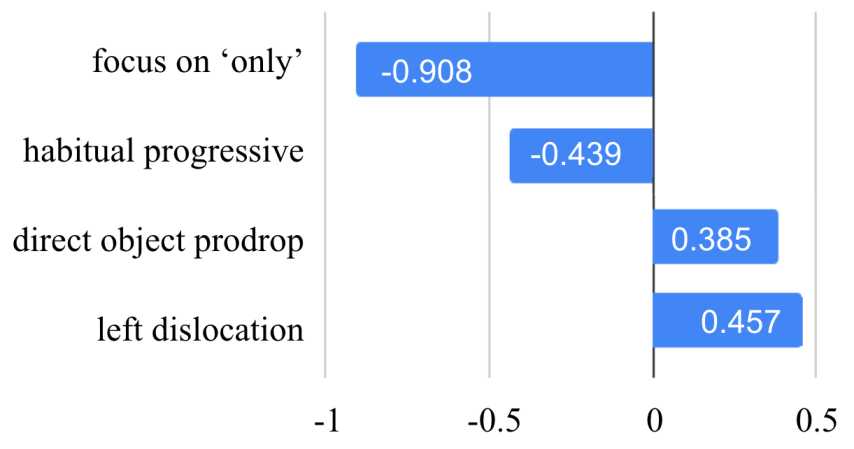}
\caption{Relative sentiment score shift showing model sensitivity to dialectal features of Indian English}
\label{fig:dialect_sentiment_hf}
\end{figure}

We find the sentiment model is sensitive to the presence/absence of dialect features. However, there was no overall trend in any one direction. Figure~\ref{fig:dialect_sentiment_hf} shows the top 2 features in terms of score shift in either direction; refer to Appendix \ref{dialect_appendix} for full results.
% As Figure~\ref{fig:dialect_sentiment_hf} shows, p
The presence of certain dialect features like \textit{left dislocation} (e.g., ``\textit{my father}, he works for a solar company'') causes a positive shift in sentiment score while other dialect features like the use of \textit{only} to signify focus (e.g., ``I was there yesterday \textit{only}'') shifts the score in the negative direction. Although it is difficult to infer systematic patterns of model behaviour due to the small number of sentences in this analysis, the high sensitivity to dialectal features prevalent in the Indian context is concerning in a fairness perspective. Finally, we note that this analysis is w.r.t to dialects of Indian vs western English. However, within India, dialects are not monolithic and resources to map dialectal features to social identities are needed to perform similar analysis for dialectal features within India.

% Presence of certain dialectal features prevelant in Indian English (e.g., \textit{direct object propdrop}, \textit{focus `only'} and \textit{left dislocation}) shifts the sentiment scores significantly, while some others such as \textit{article omission}, \textit{focus `itself'}, and \textit{preposition drop} do not change the sentiment score much. We also did not observe a general trend in the direction of shift --- i.e., while the presence of \textit{focus `only'} seemed to shift the sentiment score to the negative direction, \textit{direct object propdrop} and \textit{left dislocation} results in higher sentiment scores. 

% Although these results are w.r.t. a small number of sentences, and hence difficult to infer systematic patterns of model behaviour, the high sensitivity to dialectal features prevalent in the Indian context is still concerning from a robustness perspective.

\section{Stereotypes in Indian Context}
% \section{Fairness Evaluation}
\label{sec5_india_stereotypes}
% \begin{table}[t]

% \centering
% \begin{tabular}{|l|l|l|l|l|l|l|l|l|}
% \hline
%  \textbf{S}&
%   \textbf{=0} &
%   \textbf{\textgreater{}=1} &
%   \textbf{\textgreater{}=2} &
%   \textbf{\textgreater{}=3}  \\ \hline
% %   \textbf{\textgreater{}=4} &
% %   \textbf{\textgreater{}=5} &
% %   \textbf{\textgreater{}=6} \\ \hline
% % \textbf{Gender}     & 119  & 31  & 27  & 27 & 26 & 24 & 20 \\ \hline
% \textbf{Region}    & 2083 & 473 & 86  & 15 \\ \hline
% % & 6  & 2  & 1  \\ \hline
% \textbf{Religion}  & 692  & 604 & 229 & 52 \\ \hline
% % & 12 & 10 & 2   \\ \hline
% \end{tabular}
% \caption{Number of tuples in our dataset marked as stereotypical by 0, >=1, >=2, >=3 annotators.}
% \label{tab:num_tuples}
% \end{table}

% From the various studies presented in Section~\ref{sec4_proxies}, it is clear that predictive disparities exist along multiple axes in NLP models. Mitigating these disparities will require deeper understanding of the underlying causes. Of particular interest is 
We now turn our attention to the prevalence of social stereotypes from the Indian society in NLP data and models.
% how various social stereotypes present in the Indian society are reflected in language data and NLP models. 
There is
% A major bottleneck for such an analysis is the 
limited literature and resources on social stereotypes in the Indian context, as outlined in Section~\ref{sec2_background}. Notably, \citet{de1977regional} reported stereotypes around region and religion subgroups in India. They report the top 5 and bottom 5 traits that participants associate with 11 regional and 4 religious identities. But, the study is narrowly scoped to limited adjectives and is from decades ago thus may not reflect the current Indian society.
Recent research within NLP has built large stereotype datasets such as Stereoset \cite{stereoset-paper} and CrowS-P \cite{nangia-etal-2020-crows} to evaluate models, but they may not capture the stereotypes relevant to India. 

We build a set of stereotypical associations based on prior work but employing Indian annotators. 
%  thus curate and annotate a new dataset to that captures social stereotypes in the Indian context. 
% 
% We build on these prior work.
% to study stereotypes in Indian NLP data and models.
Like \cite{de1977regional}, we focus on the \textit{Region} and \textit{Religion}. This choice is motivated by the availability of resources and the challenges in studying the other axes (outlined in Section~\ref{sec6_discussion}).
We then use the stereotypes reported by \citet{de1977regional} and our created dataset to analyse NLP corpora and models for the prevalence of these stereotypes. 

\subsection{Dataset Creation} 
\label{dataset-creation}

We build a dataset of tuples (\textit{i}, \textit{t}) where \textit{i} is an identity term, and \textit{t} is a word token that represents a concept that is stereotypically associated (or not) with \textit{i}, for instance, (\textit{Bihari}, \textit{labourer}).
% The dataset is built in two steps: generating a list of candidate association tuples, followed by obtaining annotations on them for stereotypes from trained annotators.
 
\textbf{Generating Candidate Associations:}
We build the set of candidate association tuples (\textit{i}, \textit{t}) using identity terms described in Section~\ref{sec4_proxies} for religion and region. We then create a list of tokens based on prior work \cite{malik2021socially, nangia-etal-2020-crows, stereoset-paper}; including lists of professions, subjects of study (\textit{history}, \textit{science}, etc.), action-verbs, and adjectives for behaviour, socio-economic status, food habits, and clothing preferences. 
Tuples are formed by a cross product between tokens and identity terms. Since this cross product gives a prohibitively large number of tuples, we further prune this list by including only those tuples that co-occur (are present in the same sentence) in 
IndicCorp-en \cite{kunchukuttan2020ai4bharat} which contains 54M sentences from Indian news and magazine articles and hence likely to reflect the stereotypes prevalent in the Indian public discourse. Tuples with tokens appearing with all identity terms of a given axis are removed.

\textbf{Obtaining stereotype annotations:} 
We now obtain annotations for each tuple (\textit{i}, \textit{t}), where an annotator chooses if the association is \textit{Stereotypical} or \textit{Non-Stereotypical}. The question to the annotator was "Do you think this is a Stereotype widely held by the society?", and thus their annotations reflect community-held opinion, rather than their personal beliefs.
% Note that this stereotype may or may not be what the annotator believes or agrees with; but rather believed to be a commonly-held opinion by the general population (according to the annotator). In other words, the question was "Do you think this is a Stereotype widely held by the society?". 
% To guide this annotation, e further instructed the annotators to ask the question: \textit{Is there a societal expectation that ``Most i are t'' or “Most people who are t are i”?} and label (\textit{i},\textit{t}) as stereotypical if the answer is yes. 
They could also mark a tuple as \textit{Unsure}.
% along with an explanation.

We recruited six annotators with diverse gender and region identities: 3 male, 3 female, 2 each from the North east and Central India, and 1 each from West and South India. 
Virtual training sessions were held to explain the task with examples. We first conducted a pilot where each annotation required a justification which were reviewed by the authors, and any misconceptions were clarified. The annotators were paid 1\$ per 3 tuples.

% We select a group of 6 annotators with diverse gender and region identities, in order to effectively capture stereotypes. In particular, we select 3 male and 3 female annotators; 2 annotators belong to North-east India, 2 belong to Central India, 1 belongs to West India and 1 belongs to South India.

% Annotators are asked to mark tuples as ``Stereotypical"(S) or ``Non-Stereoypical" (NS) based on whether the candidate tuple is believed to be a stereotype (or not) in the Indian society. The annotation guidelines can be found in \ref{app-2:annotation_guidelines}

We are interested in building a ``high precision'' dataset that captures associations that are highly likely to be stereotypes held by a large portion of the society. Hence, we performed the annotation in two phases. First, each tuple is annotated by 3 annotators. The second phase is performed only for the tuples that are labeled stereotypical by at least 2 annotators in phase 1. We retain individual annotations in the dataset to capture potential differences in annotator behavior owing to their socio-cultural background and lived experiences \cite{prabhakaran2021releasing}. For the analysis presented in this paper, we report results at different levels: S\textgreater{}=1, S\textgreater{}=2, \& S\textgreater{}=3, where S denote the number of annotators who marked the tuple as stereotypical.\footnote{Too few tuples had S\textgreater{}= 4,5,6 to gain reliable insights.}
% Table~\ref{tab:num_tuples} presents the number of tuples in each bucket.
Our resource is both larger in size (See table \ref{tab:num_tuples}), and captures more diverse perspectives as compared to \citet{de1977regional}. There is only a minimal overlap (10 tuples) between the set of tuples. Table~\ref{tab:tuples_examples} shows some example tuples from our data and the number of annotators who labeled it Stereotypical.

% % Please add the following required packages to your document preamble:
% % \usepackage{booktabs}
\begin{table}[]
\small
\centering
\begin{tabular}{@{}lccccc@{}}
\toprule
      & S=0   & S\textgreater{}=1 & S\textgreater{}=2 & S\textgreater{}=3 & Total\\ \midrule
Region   & 2083 & 473              & 86               & 15 & 2556             \\
Religion & 692  & 604              & 229              & 52  & 1296             \\ \bottomrule
\end{tabular}
\caption{Number of tuples in our dataset marked as stereotypical by 0, >=1, >=2, >=3 annotators.\label{tab:num_tuples}}
\end{table}

% \usepackage{booktabs}
\begin{table}[]
\centering
\small
\resizebox{\columnwidth}{!}{
\begin{tabular}{ll}
\toprule
Tuple (identity term, attribute token) & Num. S
 \\
 \midrule
  \textbf{Region} \\
 (tamilian, mathematician) & 6 \\
 (marwari, business) & 6 \\
 (bengali, poet) & 5 \\
 (punjabi, farmer) & 4 \\
 (bihari, labourer) & 4 \\
 (bihari, farmer)& 3 \\
 (punjabi, army)& 3 \\
 (rajasthani, dance)& 3 \\
\midrule
  \textbf{Religion} \\
 (christian, missionary) & 6 \\
 (hindu, pandit)& 6 \\
 (jain, vegetarian) & 5 \\
 (muslim, butcher) & 5 \\
 buddhist, calm) & 3 \\
 (buddhist, kind) & 3 \\
 (muslim, terrorist)& 3 \\
 (sikh, angry)& 3 \\
% (tamilian,mathematician)*6 &  (christian,missionary)*6  \\
% (marwari,business)*6       & (hindu,pandit)*6          \\
% % (bengali, poet)*5        & (jain,vegetarian)*5       \\
% (punjabi,farmer)*4           & (muslim,butcher)*5        \\
% (bihari,labourer)*4        & (buddhist,calm)*3         
% % \\
% % (bihari,farmer)*3         & (buddhist,kind)*3         \\
% % (punjabi,army)*3           & (muslim,terrorist)*3      \\
% % (rajasthani,dance)*3       & (sikh,angry)*3           
\bottomrule
\end{tabular}
}
\caption{Example tuples from our dataset with number of annotators who labeled them as Stereotypical (S).\label{tab:tuples_examples}}
\end{table}

\begin{figure}[h]
\centering
\includegraphics[width=\linewidth]{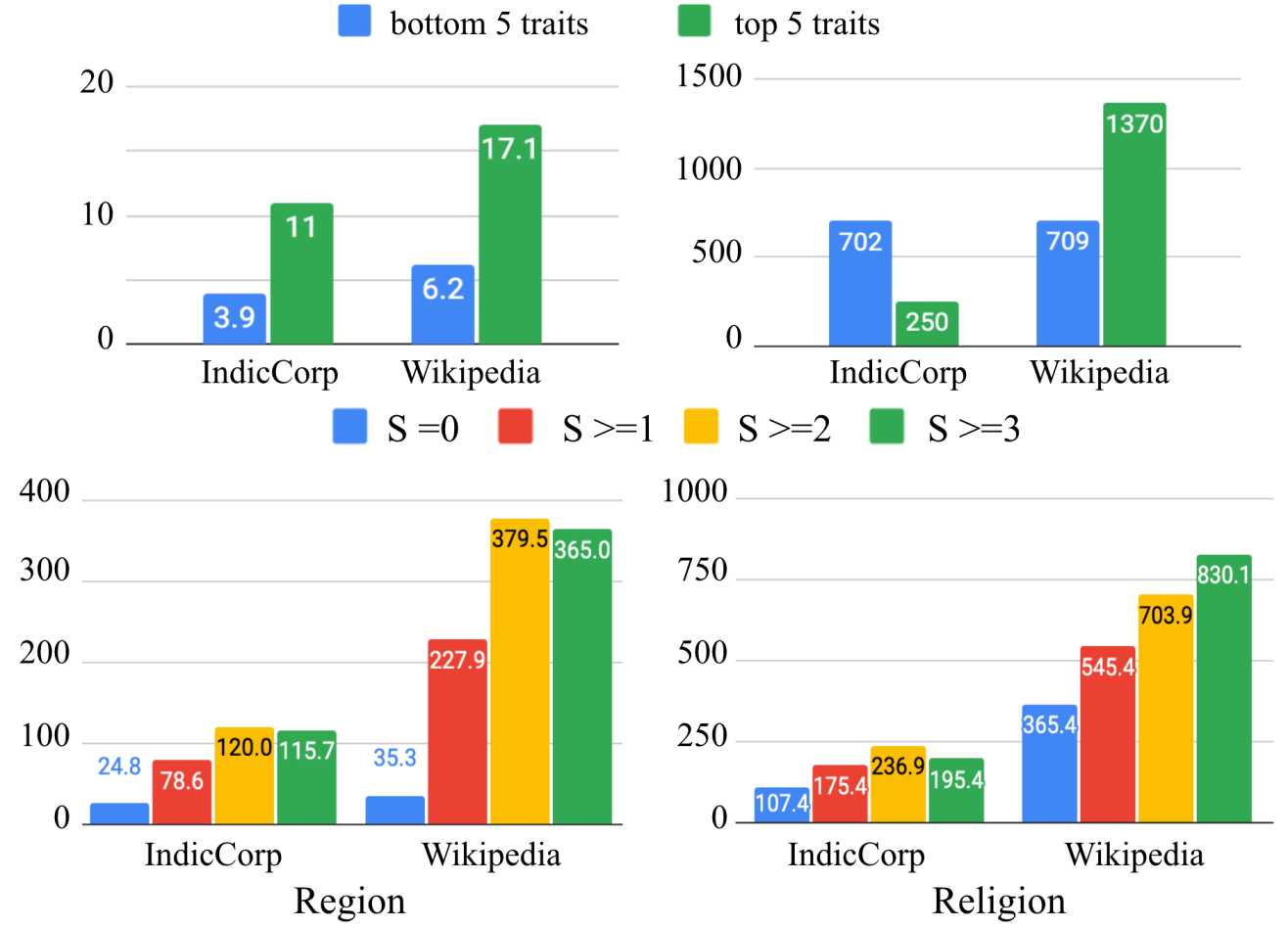}
\caption{Average co-occurrence of tuples from \citet{de1977regional} (top row) and our dataset (bottom row) in IndicCorp-en and Wikipedia
% , showing that more stereotypically associated tuples are likely to co-occur more in data.
% showing highger average co-occurrence for stereotypicall associated (top 5 traits). BottomCorpus evaluation results for tuples from \cite{de1977regional}. Top 2 graphs show average co-occurrence counts of the tuples in IndicCorp-en and Wikipedia. Bottom 2 graphs show \% of tuples in top 5 model predictions in MuRIL and mBERT.
}
% (refer \ref{model_eval} for observations)}
\label{fig:corpus_eval}
\end{figure}
\subsection{Corpus Analysis}
\label{corpus_eval}

Data can be a primary source of biases in LMs \cite{bender2021dangers}, so we analyze prevalence of stereotypical tuples in large corpora used to train LMs. We analyze the Wikipedia corpus used to train LMs like BERT \cite{bert-paper}, and the IndicCorp-en corpus used in training multilingual models like IndicBERT \cite{kakwani2020indicnlpsuite}. 
% Since IndicCorp-en contains text sourced from India, we expect it to better reflect the Indian public discourse, and hence the stereotypes, compared to Wikipedia corpus.
We measure co-occurrence counts (CC), where a tuple is considered co-occurring if both the identity term (or its plural form) and the token (or one of its inflections) occur in the same sentence.\footnote{We obtain similar trends for nPMI \cite{aka2021measuring} metric, and a window size of 2, i.e., co-occurrence within the two tokens before/after the identity term
% , obtaining similar trends
. 
% We report only the co-occurrence counts at sentence level as it is the easiest to interpret.
}

% We first analyze the tuples obtained from \citet{de1977regional}, which identifies top 5 and bottom 5 traits (adjectives) associated with 11 region and 4 religion identities. 
In the analysis using tuples from \citet{de1977regional} (Figure~\ref{fig:corpus_eval} - top row) we find co-occurrence counts are higher for tuples representing top 5 traits compared to bottom 5 traits,\footnote{One tuple for religion had very high co-occurrence in the IndicCorp-en corpus, resulting in the flipped trend.} 
% We now look at the co-occurrence counts of stereotype tuples from our data (Section~\ref{dataset-creation}); 
We observe similar trend for our dataset (Figure~\ref{fig:corpus_eval} - bottom row). Tuples that all annotators agreed to be not stereotypes (i.e., S=0) have the lowest co-occurrence counts. The average co-occurrence counts increase as more number of annotators mark the tuple as stereotype. The co-occurrence counts in Wikipedia are consistently higher, likely due its larger size as compared to IndicCorp-en (174M vs 54M sentences). In summary, we find that stereotypical associations are preferentially encoded in both corpora.

\subsection{Model Analysis}
\label{model_eval}

Following previous work \cite{filbert-paper,hutchinson2020social}, we
% follow a prompt based approach to 
probe MuRIL and mBERT 
% wherein the model 
with the task of predicting the masked token in a sentence. 
% Since our tokens belong to a handful of categories such as professions, adjectives, etc., 
We hand-craft templates for each category of tokens in our list. For e.g, a template for the profession category of tokens is: ``[\textit{i\textsubscript{t}}] are most likely to work as <MASK>."
% , where \textit{i} is replaced with the identity term of the tuple
\footnote{Complete list of templates is available with the resources.} 
For each tuple (\textit{i}, \textit{t}), we replace \textit{i\textsubscript{t}} in the template with identity term \textit{i}
% with the  probe the model with masked template sentences 
and record if the token \textit{t}, or its inflections occur in the top K (K=5)\footnote{We saw similar trends for K=3, 10, 25, 50} predictions of the model.

\begin{figure}[]
\centering
\includegraphics[width=\linewidth]{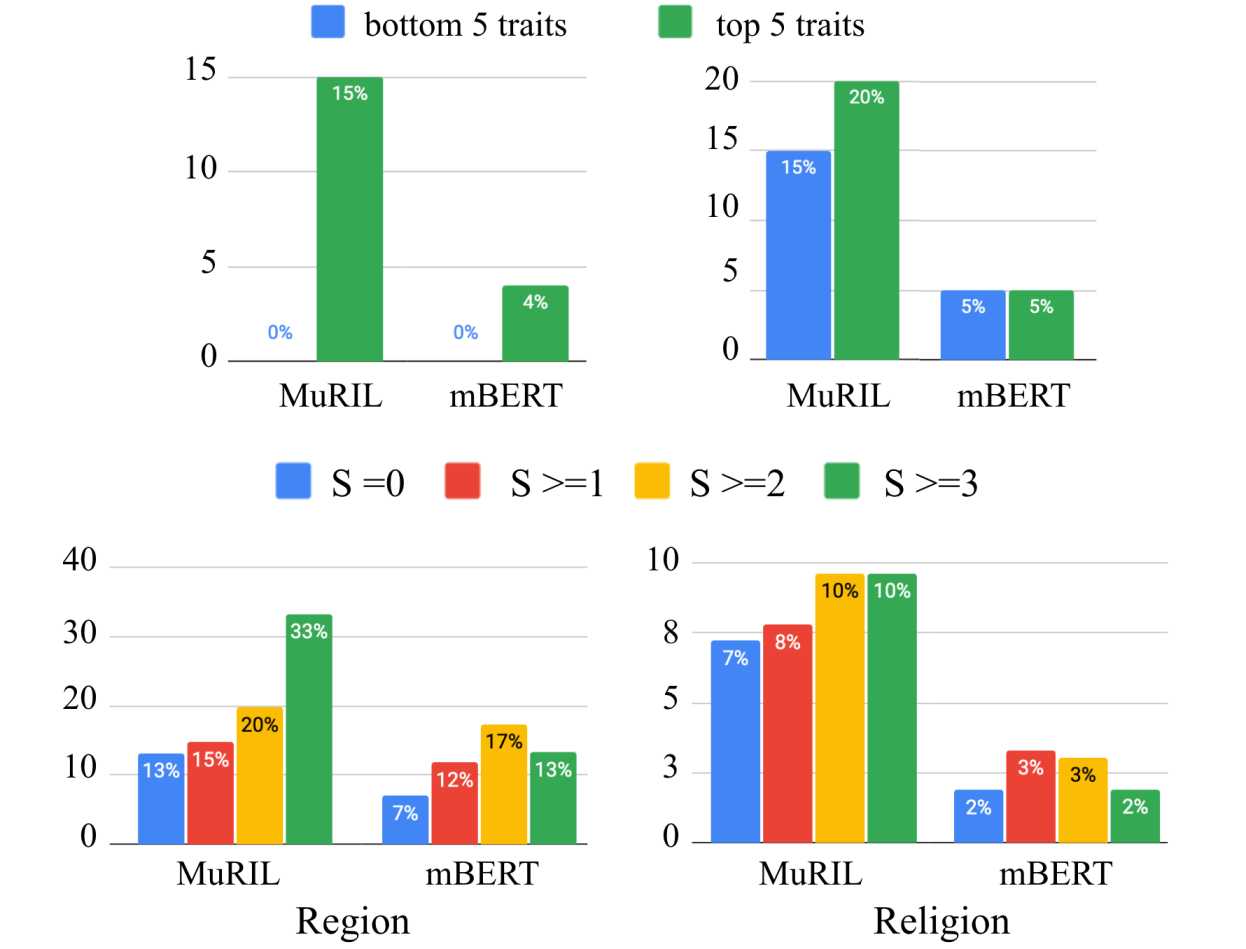}
\caption{Percentage of tuples from \citet{de1977regional} (top row) and our data (bottom row) in top 5 predictions of mBERT and MuRIL
% Corpus and model evaluation for tuples from our dataset. Top 2 graphs show average co-occurrence counts of the tuples in IndicCorp-en and Wikipedia. Bottom 2 graphs show \% of tuples in top 5 model predictions in MuRIL and mBERT.
}
\label{fig:model_eval}
\end{figure}

Figure~\ref{fig:model_eval} show the percentage of tuples occurring in top 5 predictions 
% for K=5 
for the \citet{de1977regional} and our dataset. 
% \footnote{We found similar trends for K=3, 10, 25 and 50.}
Similar to corpus analysis, for tuples from \citet{de1977regional}, we find that the top 5 associated traits 
% that are most associated with identity terms 
are more likely to appear in model predictions as compared bottom 5 traits 
% that are least associated with them, 
for both MuRIL and mBERT.
For the dataset we built, the percentage of tuples appearing in top 5 model predictions increase as more annotators label the tuple as Stereotype.\footnote{S>=3 for mBERT is an exception, with a slight dip, we leave a detailed analysis of this to future work.} 
% as we progressively restrict the set of tuples those where more number of annotators deemed them to be stereotypical. 
We also find that MuRIL shows consistently higher percentage of Stereotypical tuples in top 5 predictions
% tendency to generate stereotypical associations as compared to mBERT for both region and religion, 
suggesting that it has learned more stereotypes in the Indian context due to data sourced from India.

\subsection{Limitations}
\label{sec_limitations}

% Our dataset and the analysis presented in this section are aimed to demonstrate that popular NLP data and models do capture and propagate stereotypical associations relevant to the Indian context. 
While our dataset can serve as a starting point in evaluation and
% and serve as a starting point for future 
development of more such datasets, it is not meant as an exhaustive resource for this purpose. First of all, we capture only two axes of disparities: region and religion, and in English. We attempted to collect data for gender identity and caste, but these efforts did not yield reliable results, possibly because of the annotator pool not having the necessary familiarity with those marginalized groups and their lived experiences. 
% Future work should explore participatory approaches to provide more situated context and grounding for expanding this resource to include other axes of disparity. 
Our approach towards filtering the set of tuples for annotation based on co-occurrence limit our data to only capture those stereotypes that are explicitly mentioned in text, but there might exist stereotypes in society that are not captured in corpora and hence will not be captured by our dataset. Additionally, our methods may not capture Stereotypes that are implicit or beyond our token categories.

\section{Re-contextualizing Fairness}
% \section{Looking Forward: Towards a Holistic NLP Fairness Agenda for India}
% \section{Discussion}
\label{sec6_discussion}

\begin{figure*}[t]
\centering
\includegraphics[width=\linewidth]{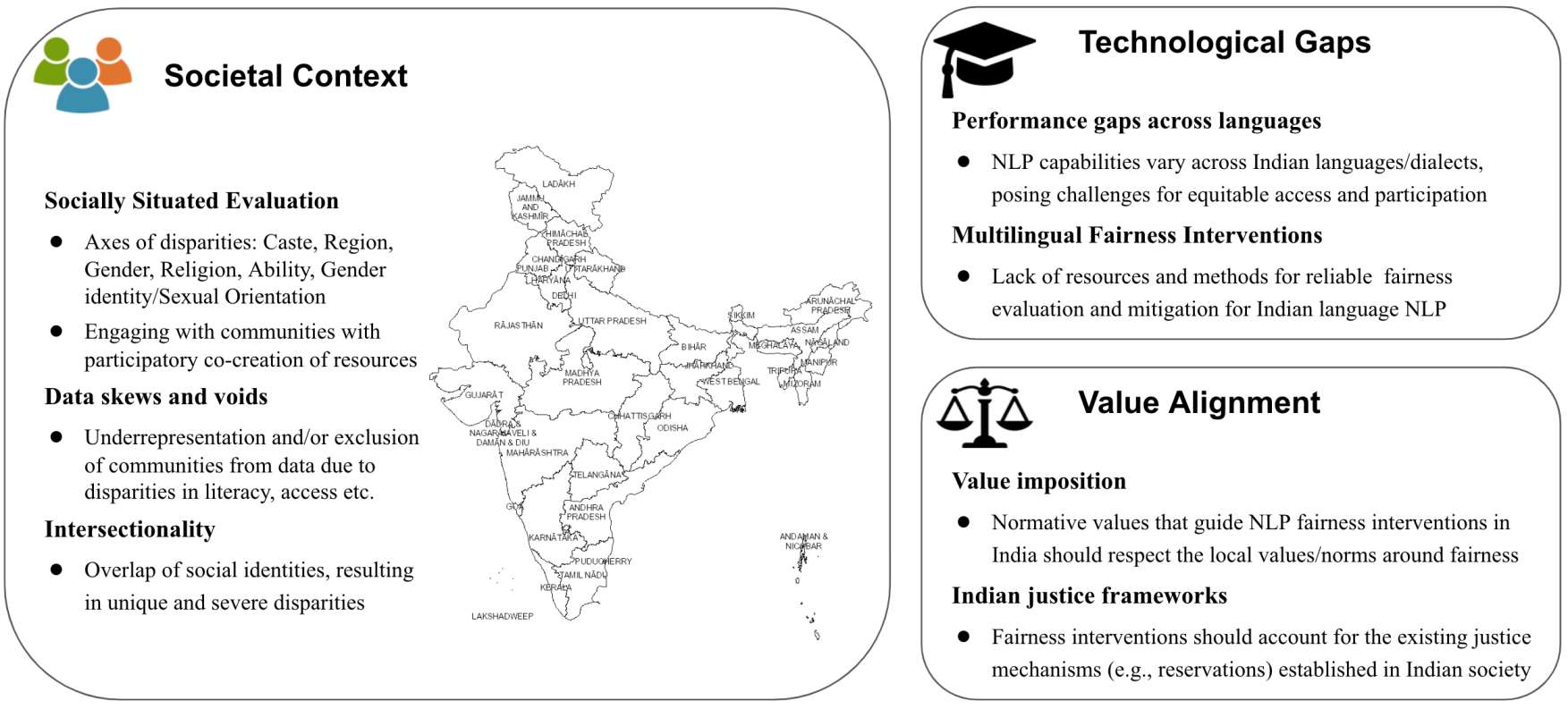}
% \caption{States and union territories of India. (Source: \url{https://indiamaps.gov.in/soiapp/}); Axes of Social Disparities and \% minority group populations; Proxies in language that \textit{may} reveal sub-group identity.
\caption{A holistic research agenda for NLP Fairness in the Indian context: accounting for societal disparities in India (Section~\ref{sec3_axes}-\ref{sec5_india_stereotypes}), bridging technological gaps in NLP capabilities/resources, and adapting fairness interventions to align with local values and norms (Section~\ref{sec6_discussion}). (Map source: \url{https://indiamaps.gov.in/soiapp/})
% \caption{Social disparities and proxies in language relevant for NLP Fairness in India. (Map source: \url{https://indiamaps.gov.in/soiapp/})
% NLP Fairness in the Indian Context. \vp{Only a placeholder figure; Feel free to design your own at \url{go/bindi-paper-figures-deck}; Objective: a pictorial representation of the agenda} \vp{Map source: \url{https://worldmapwithcountries.net/2020/03/12/india-map-with-states} to confirm copyright}
}
\label{fig:Indiaframework}
\end{figure*}

Given the empirical demonstration of biases in the Indian context in data and models, 
we now return to the broader agenda for re-contextualizing NLP fairness. We re-frame the agenda of \citet{sambasivan2019toward} 
along three aspects: accounting for \textit{Social Disparities}, bridging \textit{Technological gaps}, and adapting to \textit{Values \& Norms}. 
% We focused on the societal aspect in Sections~\ref{sec3_axes}-\ref{sec5_india_stereotypes}. 
% Based on the insights from Sections~\ref{sec3_axes}-\ref{sec5_india_stereotypes} and building on \cite{sambasivan2021re}, we now lay out a set of major challenges in accomplishing this agenda. 

\subsection{Accounting for Indian \textit{Societal} context}

We provided a comprehensive account of prominent axes of disparities in Indian society (Section~\ref{sec3_axes}), and demonstrated biases around them encoded in NLP data and models (Section~\ref{sec4_proxies}-\ref{sec5_india_stereotypes}). Our work is just the first step and is far from over. 
% We outline some of the limitations of our work, and outline major challenges going forward. 

\paragraph{Socially Situated Evaluation:}
Most of our analysis is focused on region and religion. A major hurdle in expanding axis coverage is the (lack of easy) access to diverse annotator pools who have familiarity and/or lived experiences of the marginalized groups especially as the public discourse around (dis)ability, gender identity and sexual orientation is relatively new and limited. 
We believe that participatory approaches \cite{lee2020human} to create resources for fairness evaluation will be crucial for meaningfully addressing this gap.
% We believe that participatory approaches such as the Masakhane effort are crucial for addressing this gap in Fairness research.\footnote{\url{https://www.masakhane.io/}}

\paragraph{Data Voids:}
Social disparities in literacy and internet access might cause entire communities to be excluded from language data \cite{sambasivan2021re}. 
% While recent efforts have initiated building LMs using data sourced from India \cite{khanuja2021muril}, 
% Not accounting for such data voids might result in representational harms \cite{blodgett2020language} being baked into the LMs.
% that has become base infrastructure for NLP
% \cite{bommasani2021opportunities}.   
Further, the risk of unintentionally excluding marginalized communities based on dialect or other linguistic features while filtering data to ensure quality \cite{dodge2021documenting,gururangan2022whose} is even higher in the Indian context because of very limited computational representation of marginalized communities. 
Accounting for data voids and intentional data curation (such as by collecting language data specifically from marginalized communities \cite{abraham-etal-2020-crowdsourcing, nekoto-etal-2020-participatory}) can significantly help bridge this gap.
% Participatory approaches to data curation such as that done by \citet{abraham-etal-2020-crowdsourcing} to collect speech data from tribal and rural Maharashtra can significantly help bridge such data voids.

% While large LMs have become standard infrastructure for NLP research and development \cite{bommasani2021opportunities,bender2021dangers}, they are largely trained on data sourced from the West. For instance, \citet{johnson2022ghost} point out that popular LMs such as GPT-3 exhibit dominant US values, when analyzed with value sensitive topics. While recent efforts have initiated work on building large LMs with data from Indian contexts \cite{khanuja2021muril}, these are prone to misrepresentation and exclusion of entire communities \cite{sambasivan2019toward}, who are typically already marginalized. Such data typically come from internet sources like social media, Wikipedia, books, and news corpora \cite{kakwani2020indicnlpsuite, khanuja2021muril}; however, internet participation is limited to those with education, access to technology, and digital awareness. Further, the risk of unintentionally excluding marginalized communities based on dialect or other linguistic features while filtering data to ensure quality \cite{dodge2021documenting,gururangan2022whose} is even higher in the Indian context because of very limited computational representation of marginalized communities. Thus, data voids should be taken into consideration while building data resources and models.

\paragraph{Intersectionality:}
% The overlap of various social identities, such as religion, gender, ability, etc,  contributes to the specific type of systemic oppression and discrimination experienced by an individuals or subgroups \cite{collins2020intersectionality}.
Due to the interplay of all the diverse axes in the Indian context, intersectional biases \cite{collins2020intersectionality} experienced by different marginalized groups are often more severe \cite{dalit-women-in-india}. With notable differences in literacy, economic stability, technology access, and healthcare access across geographical, caste, religious, and gender divides, representation in and access to language technologies are also disparate. Bias evaluation and mitigation interventions should account for these intersectional biases.
% by marginalised communities.

% \paragraph{Participatory Design}

% The purpose of the dataset we create is to serve as a first step towards evaluating bias in models and data in India specific context. However, having a limited number of annotators, this may not capture pluralistic nuances. Further, without involvement of the communities that face marginalization, there is a risk of not capturing their lived experiences. For example, in our dataset, one of the gender sub-groups is "transgender", however none of the annotators we have identify as transgender (to the best of our knowledge). This limits the capability of typical NLP style annotation practices (such as this one) to capture the lived experiences of marginalized communities. More so because data workers employed by annotation firms and crowd-sourcing platforms tend to be urban, educated, and digitally-literate \cite{abraham-etal-2020-crowdsourcing} who represent a elite fraction of the Indian society. To avoid creating datasets that have a myopic view point, especially in inherently subjective tasks like Stereotype annotations, participatory design with active involvement of communities on the ground is necessary to create reliable resources reflect the ground reality of marginalization.

\subsection{Bridging cross-lingual \textit{Technological} gaps}

While we focus on English language data and models in this paper, it is crucial to mitigate the gaps in NLP capabilities and resources across Indian languages, both in general and for fairness research. 

\paragraph{Performance gaps across languages:}
% \paragraph{Fair Performance across Languages:}
India is a vastly multilingual country with hundreds of languages and thousands of dialects. But there are wide disparities in NLP capabilities across these languages and dialects. These disparities pose a major challenge for equitable access, creating barriers to internet participation, information access, and in turn, representation in data and models. While the Indian NLP community has made major strides in addressing this gap in recent years \cite{khanuja2021muril}, more work is needed in building and improving NLP technologies for marginalized and endangered languages and dialects.

% Although India is a vastly multilingual country with 22 recognised languages, and thousands of varieties and dialects, there are wide disparities in NLP technology capabilities across these languages and dialects. While more work needs to be done to assess and mitigate effects of unfair societal biases in NLP in Indian languages, disparate performance across these languages pose fairness challenges regardless of those biases. Poor technology in low-resource languages can in turn create barriers to internet participation, information access, and cyclically, representation in data and models.  It is thus essential that the performance of language technologies is not widely disparate for various (marginalised) sub-groups preventing them the opportunity to access and utilize technology effectively

\paragraph{Multilingual fairness research:}
NLP Fairness research relies on bias evaluation resources 
% that are built in the Western context. 
and while we present such resources for the Indian context, we limited our focus to only English. It is crucial to expand this effort into Indian languages, along the lines of recent work on Hindi, Bengali, and Telugu \cite{malik2021socially,pujari2019debiasing}. This is especially important since biases may manifest differently in data and models for different languages. Additionally, how bias transfers in transfer-learning paradigms for multilingual NLP is unknown. Finally, bias mitigation in one (or a few) language(s) may have counter-productive effects on other languages. 
Hence, a research agenda for fair NLP in India should address these various unknowns that the dimension of language brings.

\subsection{Adapting to Indian \textit{Values and Norms}}

Fairness interventions essentially impart a normative value system on model behaviour. It is crucial to ensure that these interventions are not at odd with Indian values, norms, and legal frameworks.

% \paragraph{Philosophical Roots}

\paragraph{Accounting for Indian justice models:}
India has established legal restorative
justice measures for resource allocation, colloquially known as the ``reservation system'' \cite{ambedkar2014annihilation}, where historically marginalized communities (like Dalits, backward castes, tribals, and religious minorities) are afforded fixed quotas in educational and government institutions to counter historical deprivation. NLP fairness interventions should conform to these 
% on these domains should also consider how they work in the context of such 
established measures that are otherwise non-existent, and hence not thought for in the West.

\paragraph{Avoiding value imposition:}
Fairness inquiries answer questions such as: what fairness means, and how fair is fair enough? These questions, and their answers risk value imposition. While, implicitly these answers draw largely from Western values rooted in {egalitarianism, consequentialism, deontic justice, and Rawls’ distributive justice} \cite{sambasivan2021re}, the philosophy of fairness in India is rooted in social restorative justice. More work should look into such value alignment challenges for 
% which is not trivial when it comes to deploying 
fairness interventions \cite{gabriel2020artificial}.

\section{Conclusion}
In this paper, we holistically re-contextualize fairness research for the Indian context taking an NLP-centric lens to \citet{sambasivan2021re}. 
% and exwe take a NLP-centric lens to extend work of \citet{sambasivan2019toward} for recontextualiz a comprehensive research agenda to re-contextualize fairness research in NLP for the Indian context. Building on the foundational work by \citet{sambasivan2021re}, 
We lay out a research agenda advocating to account for the societal context in India, bridge technological gaps in capability and resources, and align with local values and norms (Section~\ref{sec6_discussion}). Our focus here is on India, but the broader framework of this work can be used to recontextualize fairness for any geo-cultural context. We
% delve deeper into the societal disparities in this paper, 
outline the prominent axes of disparities in India (Section~\ref{sec3_axes}), and demonstrate biases around them in NLP models and corpora. To summarize: First, our perturbation analysis reveals that sentiment model predictions are significantly sensitive to regional, religious, and caste identities (Section~\ref{identity_terms}), and dialectal features (Section~\ref{dialect}). Second, our DisCo analysis shows the necessity of India-specific resources for revealing biases in the Indian context (Section~\ref{personal_names}). Third, we build a stereotype dataset for the Indian context and demonstrate preferential encoding of stereotypical associations in both NLP data and models (Section~\ref{sec5_india_stereotypes}). 
% Across our analyses, we find that models trained specifically using Indian data (i.e., MuRIL) are more likely to pick up Indian biases and stereotypes. 
% We will release our resources for reproducibility and to aid future work.\footnote{\url{tinyurl.com/anon-india-fairness-data}} 
While there is more work to be done, we believe this is an essential first step towards a meaningful NLP fairness research agenda for India.

\section{Ethical considerations}
We build resources to demonstrate biases in models, these resources alone are insufficient to capture all the undesirable biases in the Indian society. As described in Section~\ref{sec_limitations}, our dataset lacks coverage across the various Indian axes of disparities, languages, and reflects the judgements of a small number of annotators. Hence, they should be used only for diagnostic and research purposes, and not as benchmarks to prove lack of bias. We also urge that the list of names with prototypical binary gender associations from Wikipedia (used in Section \ref{personal_names}) not be used to train gender prediction models.

\section*{Acknowledgements}
We thank Nithya Sambasivan for her groundbreaking research and early guidance on this project. We thank Ben Hutchinson, Kellie Webster, Ding Wang, Molly FitzMorris, and Reena Jana for their critical insights on earlier drafts. We are grateful to the anonymous reviewers for their helpful feedback. We thank Dinesh Tewari for his work on facilitating the project. We thank the annotation team for facilitating our data collection.

% \section{Ethical Considerations}
% \input{Re-contextualizing_Fairness_in_NLP_The_Case_of_India/V3_aacl/sec8_ethical}
% \label{sec_ethical}

\bibliography{0-main.bib}
\bibliographystyle{acl_natbib}

\newpage
\appendix
\section{Perturbation Sensitivity Analysis with dialectal features: full results}
\label{dialect_appendix}
In \S\ref{dialect} we perform perturbation sensitivity analysis with sentences from \citet{demszky2021learning}. Here we provide the complete results for this analysis, where in-text we provided only the top-2 most positively shifted and negatively shifted features.

\begin{table}[h]
\centering
\small
\begin{tabular}{ll}
\toprule \\
\textbf{Dialectal Feature} & \textbf{Relative sentiment} \\
& \textbf{score shift} \\
\midrule \\
focus `only' &	-0.908 \\
habitual progressive &	-0.439 \\
inversion in embedded clause &	-0.412 \\
topicalized non-argument constituent &	-0.205 \\
lack of copula &	-0.029 \\
stative progressive &	-0.019 \\
invariant tag ('isn't it', 'no', 'na') &	-0.010 \\
focus 'itself' &	-0.007 \\
resumptive object pronoun &	0.000 \\
non-initial existential 'X is / are there' &	0.004 \\
resumptive subject pronoun &	0.009 \\
mass nouns as count nouns &	0.009 \\
article omission &	0.023 \\
preposition drop &	0.025 \\
lack of inversion in wh-questions &	0.036 \\
extraneous 'the' (often generic) or 'a' &	0.084 \\
prepositional phrase fronting &	0.186 \\
object fronting &	0.192 \\
use of 'and all' &	0.208 \\
lack of agreement &	0.274 \\
direct object prodrop &	0.385 \\
left dislocation &	0.457 \\
\bottomrule
\end{tabular}

\caption{Relative sentiment score shift due to presence or absence of dialectal features}
\end{table}

\end{document}